\documentclass[prepriMPnt,12pt]{elsarticle}

\usepackage{graphicx}
\usepackage{afterpage}
\usepackage{geometry}
\usepackage{ulem}
\usepackage{amstext}
\usepackage{amsmath,amssymb,amsfonts}
\usepackage{url}
\usepackage{subcaption}
\usepackage{float}
\usepackage{changepage}

\usepackage{graphicx}

\newcommand{\Dm}{\mathbf{D}}
\newcommand{\Am}{\mathbf{A}}
\newcommand{\Bm}{\mathbf{B}}
\newcommand{\vb}{\mathbf{v}}
\newcommand{\wb}{\mathbf{w}}
\newcommand{\qb}{\mathbf{q}}

\newcommand{\condin}{\perp \!\!\! \perp}

\journal{arXiv}

\begin{document}

\begin{frontmatter}



\title{Learning Multimorbidity Patterns from Electronic Health Records Using Non-negative Matrix Factorisation}


\author[dm,tgi,brc]{Abdelaali Hassaine}
\author[dm,tgi,brc,australia]{Dexter Canoy}
\author[dm,tgi,brc]{Jose Roberto Ayala Solares}
\author[dm,tgi]{Yajie Zhu}
\author[dm,tgi]{Shishir Rao}
\author[dm,tgi]{Yikuan Li}
\author[dm,tgi,brc]{Mariagrazia Zottoli}
\author[dm,tgi,brc]{Kazem Rahimi}%\corref{cor1}}
\author[dm,tgi]{Gholamreza Salimi-Khorshidi}

\address[dm]{Deep Medicine, Oxford Martin School, University of Oxford, Oxford, United Kingdom}
\address[tgi]{The George Institute for Global Health (UK), University of Oxford, Oxford, United Kingdom}
\address[brc]{NIHR Oxford Biomedical Research Centre, Oxford University Hospitals NHS Foundation Trust, Oxford, United Kingdom}
\address[australia]{Faculty of Medicine, University of New South Wales, Sydney, Australia}           
%\cortext[cor1]{Corresponding author: kazem.rahimi@georgeinstitute.ox.ac.uk}

\begin{abstract}
Multimorbidity, or the presence of several medical conditions in the same individual, has been increasing in the population, both in absolute and relative terms. However, multimorbidity remains poorly understood, and the evidence from existing research to describe its burden, determinants and consequences has been limited. Previous studies attempting to understand multimorbidity patterns are often cross-sectional and do not explicitly account for multimorbidity patterns' evolution over time; some of them are based on small datasets and/or use arbitrary and narrow age ranges; and those that employed advanced models, usually lack appropriate benchmarking and validations. In this study, we (1) introduce a novel approach for using Non-negative Matrix Factorisation (NMF) for temporal phenotyping (i.e., simultaneously mining disease clusters and their trajectories); (2) provide quantitative metrics for the evaluation of disease clusters from such studies; and (3) demonstrate how the temporal characteristics of the disease clusters that result from our model can help mine multimorbidity networks and generate new hypotheses for the emergence of various multimorbidity patterns over time. We trained and evaluated our models on one of the world's largest electronic health records (EHR), with 7 million patients, from which over 2 million where relevant to this study.
\end{abstract}

%\includegraphics[width=1\linewidth,keepaspectratio=true]{./graphical_abstract.pdf}

%\item Despite the increasing prevalence of multimorbidity (i.e., the presence of several medical conditions in the same individual) in the population, its patterns and trajectories remain poorly understood. 
%\item In this paper, we introduce a new approach for temporal phenotyping -- using non-negative matrix factorisation (NMF) -- and provide a solution for benchmarking/evaluating the resulting disease clusters and multimorbidity networks that are derived from these disease clusters' time courses.
%\item Our approach, which is validated on an EHR dataset of 7 million patients, can provide the field of multimorbidity research with new solutions to investigate the patterns by which diseases occur over time in the population and in one's life.


\end{frontmatter}


\section{Introduction}
\label{intro}
Multimorbidity is generally defined as the presence of two or more chronic conditions in an individual~\cite{van1998multimorbidity}. There is growing evidence that the number of people with multimorbidity has been increasing in many populations, both in relative and absolute terms. This increasing burden has been attributed to a number of factors, including the trend towards an ageing population demographics, as well as factors relating to changes in lifestyle, health-seeking behaviour and the environment~\cite{tran2018patterns}. Research in this area has been growing, but most investigations have focused on predicting, preventing and managing disorders in isolation from one another. Therefore, more research is needed for a better understanding of this emerging burden, its underlying patterns and mechanisms, in order to anticipate its consequences for the health services and the provision of appropriate care~\cite{ams2018}.

Different types of studies in the past have tried several methods to investigate multimorbidity. We refer to the first group of such studies as ``pairwise methods''; in these methods, disease pairs that show co-occurrence frequencies that are different from what their individual frequencies in the population would predict, are considered to be ``connected''~\cite{goldacre2000use, hidalgo2009dynamic, jensen2014temporal, strauss2014distinct}. Treating diseases as nodes and connectedness as edges, these studies formed networks of which the properties were then used to characterise various multimorbidity pathways. Liu et al.~\cite{liu2015temporal} took the idea further and formed a network where nodes were medications, tests and diagnoses. While pairwise methods are valuable in generating comorbidity hypotheses for disease pairs, their inability to address conditional independence (i.e., where the correspondence between disease $c_i$ and $c_j$ is due to a disease $c_k$ they both are linked to, i.e., $P(c_i|c_j,c_k)=P(c_i|c_k)$, or $c_i \condin  c_j~|~c_k$~\cite{pearl2009causality}) can make the multi-disease networks resulting from them misleading.

This has led to the rise of alternative methods for disease phenotyping and the study of multimorbidity, that can deal with multiple diseases simultaneously. Except a limited number of studies that used ``probabilistic methods'' such as latent class growth analysis~\cite{strauss2014distinct} and Hidden Markov Model~\cite{wang2014unsupervised}, majority of recent studies in the field have relied on ``factorisation methods''. The earlier factorisation methods in this field started by forming a matrix, where the entry $i,j$ denotes a metric related to disease $i$ (or other concepts in EHR, such as medication and/or clinical measurements) in patient $j$; factor methods decompose such a matrix into $P$ multimorbidity patterns (MPs), each consisting of a disease cluster (DC) and the expression of that DC across patients. While relatively effective, such matrix factorisation approaches did not take into account the temporal aspect of MPs (i.e., disease-based temporal phenotyping) and have been mostly limited to mining static disease (and sometimes medication) clusters~\cite{holden2011patterns,schafer2010multimorbidity,marengoni2009patterns,kirchberger2012patterns,roso2018comparative}. As an exception, Zhou et al.~\cite{zhou2014micro} introduced a matrix factorisation method that considers the temporal patterns in EHR data; while the method can be employed for multimorbidity analyses, the study's primary focus was on the prediction of future diseases (as opposed to multimorbidity). 

On the other hand, solutions for joint phenotyping (based on diseases plus other concepts, for instance) have been achieved through the use of tensor factorisation. For instance, both~\cite{ho2014marble} and~\cite{wang2015rubik}, used non-negative tensor factorisation for joint phenotyping of diagnoses and medications, through adding medications as a new dimension to the input matrix (and hence the tensor). Following a similar approach Perros et al. \cite{perros2019temporal}, introduced a time dimension and used tensor factorisation for joint phenotyping of diseases and time (i.e., disease-based temporal phenotyping); instead of explicitly dealing with time, however, their model deals with the chronological order/index of encounters. In another recent study of temporal phenotyping, Zhao et al.~\cite{zhao2019detecting} used tensor factorisation and showed the effectiveness of their resulting phenotypes for stratifying the risk of cardiovascular diseases. %[AH] I still think we need to criticise this study as well, otherwise , the reviewer may ask how is our approach was different. I think the fact they considered the time prior to cariovascular disease is a flaw which makes their temportal phenotypes specific to CVDs and not necessarily generalisable.
% [RK] what do you mean? I don't think I follow.
In summary, previous temporal phenotyping studies relied on relatively complex tensor-based factorisation approaches, which makes the use of simpler matrix factorisation techniques for this goal under-explored.

In this study, we introduce a novel design that can enable matrix factorisation techniques, such as NMF~\cite{cichocki2009fast,fevotte2011algorithms}, for disease-based temporal phenotyping (i.e., to mine the DCs and their expression over time in patients). This was facilitated by an assumption that is key to our study: the underlying DCs are the same for all patients, but their expression pattern (i.e., the strength of a DC's expression in a given year of one's life) varies from one person to the other. We trained our approach on one of the world's largest EHR datasets -- consisting of more than 7 million individuals (2 million were appropriate for this study) -- and evaluated its results for the study of multimorbidity. Given that the past studies of multimorbidity are often cross-sectional, use arbitrary narrow age ranges, and lack appropriate benchmarking and validations (of their methodology and results), this paper can introduce a new methodology pipeline to the field of multimorbidity research that can employ the temporal information -- hidden in longitudinal datasets such as EHR -- to generate and evaluate new hypotheses on how diseases occur over time.

\section{Materials and methods}
\label{sec:materials_and_methods}
In this section, we explain the source of our EHR data, our approach in using NMF, and the network analyses that we carried out to show the strength of temporal disease-based phenotyping for the study of MPs. Note that, for the rest of the paper, matrices will be denoted by upper case bold fonts (e.g., $\Am$), vectors will be denoted by lower case bold fonts (e.g., $\mathbf{a}$), and everything else (scalar and indices) will be denoted with no bolding of the fonts.

\subsection{EHR Data}
An important factor that can help address the aforementioned issues in the current state of multimorbidity studies, is the rapid growth in the development of healthcare information systems, and the growing interest in utilising EHR. Particularly, the longitudinal nature of EHR and its richness (e.g., containing diagnoses, medications, and tests/measurement) can provide a unique opportunity to study temporal multi-modal phenotyping. 

In this study, we used the Clinical Practice Research Datalink (CPRD)~\cite{cprd2015}; its longitudinal primary care data from a network of 674 general practices (GP)
in the UK, and is linked to secondary care (i.e., Hospital Episode Statistics, HES) and other health and administrative databases (e.g., Office for National Statistics' death registration). One in ten GPs in the UK contributes data to CPRD; it covers 35 million patients, among whom nearly 7 million currently registered patients.
CPRD is broadly representative of the population by age, sex, and ethnicity~\cite{herrett2015data}. It has been extensively validated and is considered as the most comprehensive longitudinal primary care database~\cite{walley1997uk}, with several large-scale epidemiological reports~\cite{emdin2015usual,emdin2016usual,smeeth2004risk} adding to its credibility. 

HES, on the other hand, contains data on hospitalisations, outpatient visits, accident and emergency for all admissions to National Health Service (NHS) hospitals in England~\cite{lee2002top}. Approximately 75\% of the CPRD GPs in England (58\% of all UK CPRD GPs) participate in patient-level record linkage with HES, which is performed by the Health and Social Care Information Centre~\cite{mohseni2017influenza}. In this study, we only considered the data from GPs that consented to (and hence have) record linkage with HES. The importance of primary care at the centre of the National Health System in the UK, the additional linkages, and all the aforementioned properties, make CPRD one of the most suitable EHR datasets in the world for data-driven clinical/medical discovery and machine learning.

\subsection{Study Population}
In order to have a comprehensive coverage of patients' health journey, we only considered patients with at least 5 years of follow-up; this resulted in a total number of 2,204,178 patients with a total number of 25,791,493 clinical encounters. Note that, in this study, we are interested in ``incident cases'' and not in the ``prevalent cases'' of diseases, i.e. new occurrences of diseases rather than diseases carried over. Therefore, we only considered the first occurrence of each disease happening after 1 year of any patient's registration date with the general practice clinic (as the first year after registration is likely to contain diseases carried over rather than new occurrences of diseases).

Another important step in processing CPRD was to create consistent disease classifications between GP and HES data and choosing the appropriate level of granularity in diseases' hierarchy
%(see Table~\ref{tab:icd_hierarchy})
. In HES, diseases are coded using ICD-10 (International Classification of Diseases~\cite{icd10_2016}), whereas in the GP records diseases are coded using Read Code~\cite{read_codes}. ICD-10 offers a hierarchical form that makes its use for data mining and machine learning much more convenient. Therefore, we decided to map the diseases to the ICD-10 domain, i.e., we mapped the Read Codes to ICD-10 codes using the mapping provided by NHS Digital~\cite{read_icd10_cross_map}. When no direct mapping was available, Read Codes were first mapped to SNOMED-CT codes~\cite{snomed_codes} (also provided by NHS Digital~\cite{read_snomed}), and the latter are then mapped to ICD-10 codes using the mapping provided by the US National Library of Medicine~\cite{snomed_icd10}.

\iffalse
\begin{table}
\caption{The number of diseases in each level of ICD-10 hierarchy}
\label{tab:icd_hierarchy}
\begin{tabular}{|l|l|l|}
\hline
\textbf{Level} & \textbf{Count} & \textbf{Example} \\\hline
ICD-10 chapter & 22 & II Neoplasms \\\hline
ICD-10 block & 211 & C00-C97 Malignant neoplasms \\\hline
ICD-10 2 digits level & 2049 & C00 Malignant neoplasm of lip \\\hline
ICD-10 full code & 10138 & C00.0 Malignant neoplasm: External upper lip \\\hline
\end{tabular}
\end{table}
\fi

\begin{figure}[H]
\begin{center}
	\includegraphics[width=1\linewidth,keepaspectratio=true]{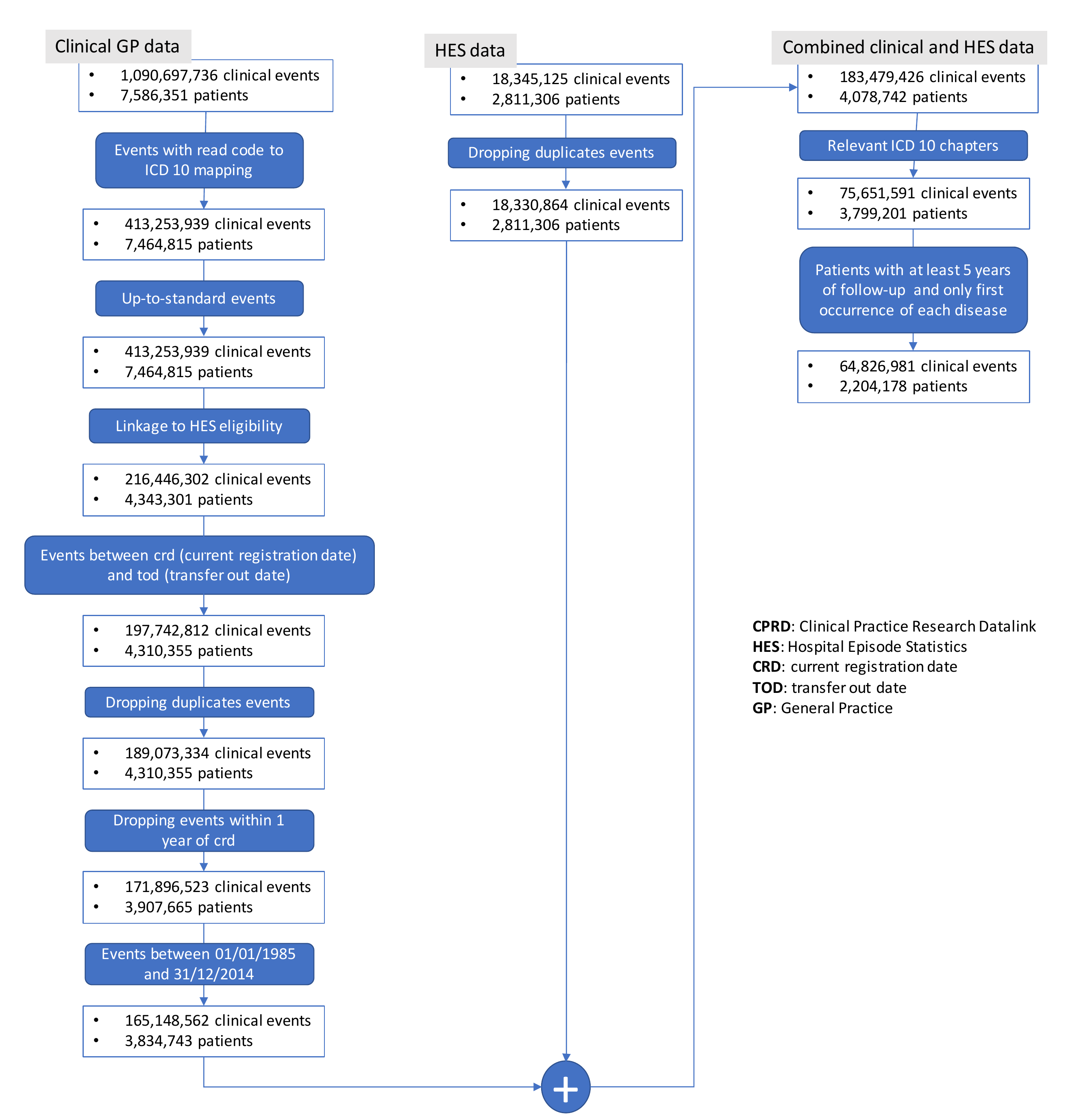}
\end{center}
\caption{Inclusion/exclusion criteria for the study. In order to make sure that the data is appropriate for the study's objectives (described at the end of Section~\ref{intro}), we made sure that only patients who meet our criteria (in terms of follow up, diseases of interest, and quality of linkage) are included. Also note that, terms such as ``up to standard'', ``crd'' and ``tod'' are referring to fields in CPRD data; for more details on these, we refer readers to CPRD manual~\cite{cprd2015}.}
\label{fig:datacut_flowchart}
\end{figure}

As mentioned earlier, ICD-10 codes are organised in a tree-like hierarchy. Working at the highest level of the ICD-10 hierarchy will result in only a few diseases and hence is not likely to help unravel the complex underlying dependencies among diseases. Conversely, working at the lowest level of the hierarchy will generate thousands of diseases, each with small number of occurrences (and even a smaller number of co-occurrences) and hence an overall difficulty of mining useful inter-disease patterns. In this study, similarly to the work in~\cite{roso2018comparative}, we chose to work at the ICD-10 block level, which provides a good trade off between granularity and co-occurrence, and can lead to medically interpretable results. Furthermore, for the purpose of this study, we eliminate all the diagnoses relating to pregnancy, general symptoms, external causes and administration (i.e., ICD-10 chapters XV, XVI, XVIII, XIX, XX and XXI); similar to~\cite{jensen2014temporal}. The resulting 142 ICD-10 blocks are what we will refer to as diseases from here on -- more details on our inclusion/exclusion process is shown in Figure~\ref{fig:datacut_flowchart}.

\begin{figure}[H]
	\centering	
	\begin{subfigure}{.49\linewidth}
	   \includegraphics[width=\textwidth,height=0.15\paperheight,keepaspectratio=true]{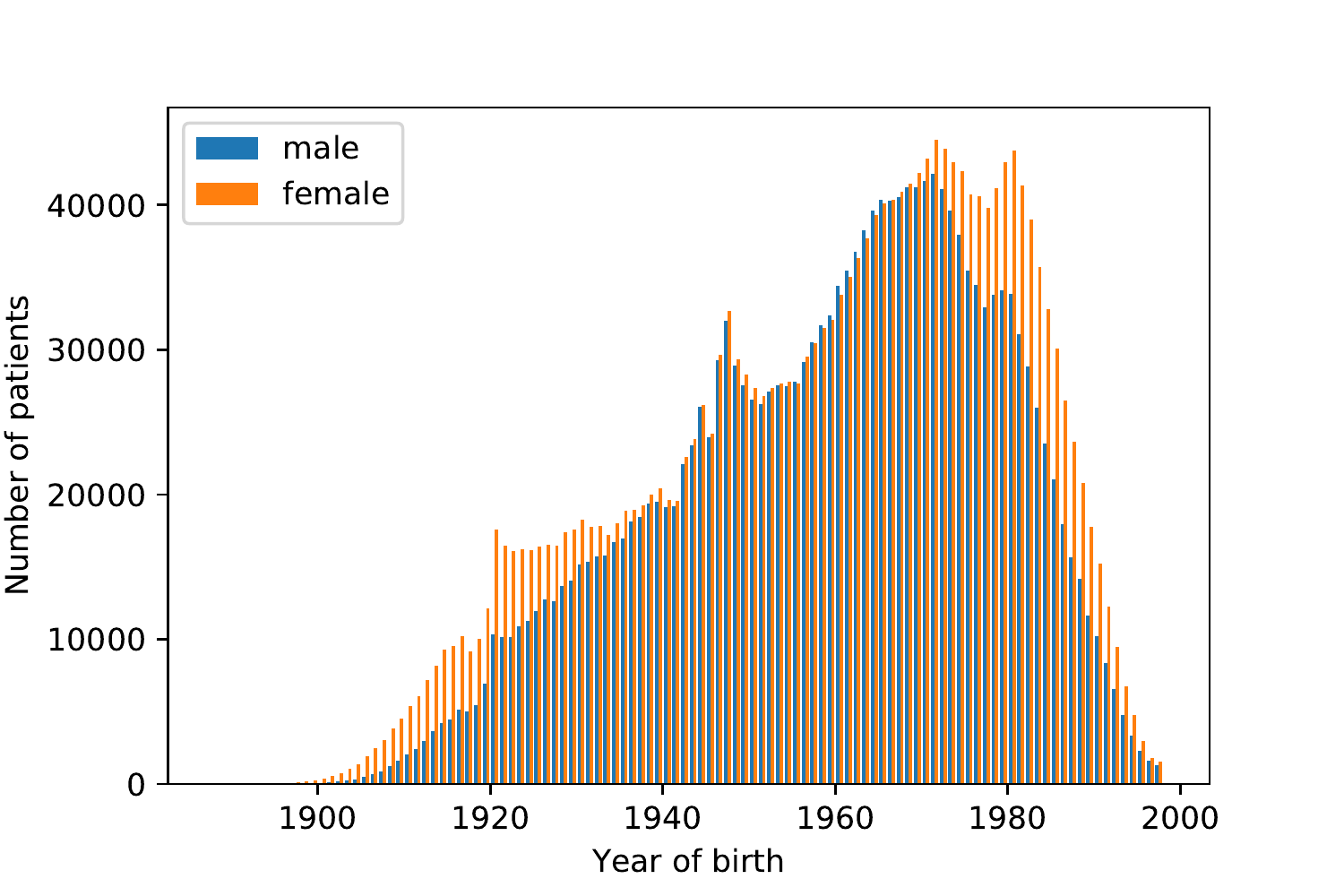}
	   \caption{Distribution of the patients' birth year}
	\end{subfigure}	
	\begin{subfigure}{.49\linewidth}
			\includegraphics[width=\textwidth,height=0.15\paperheight,keepaspectratio=true]{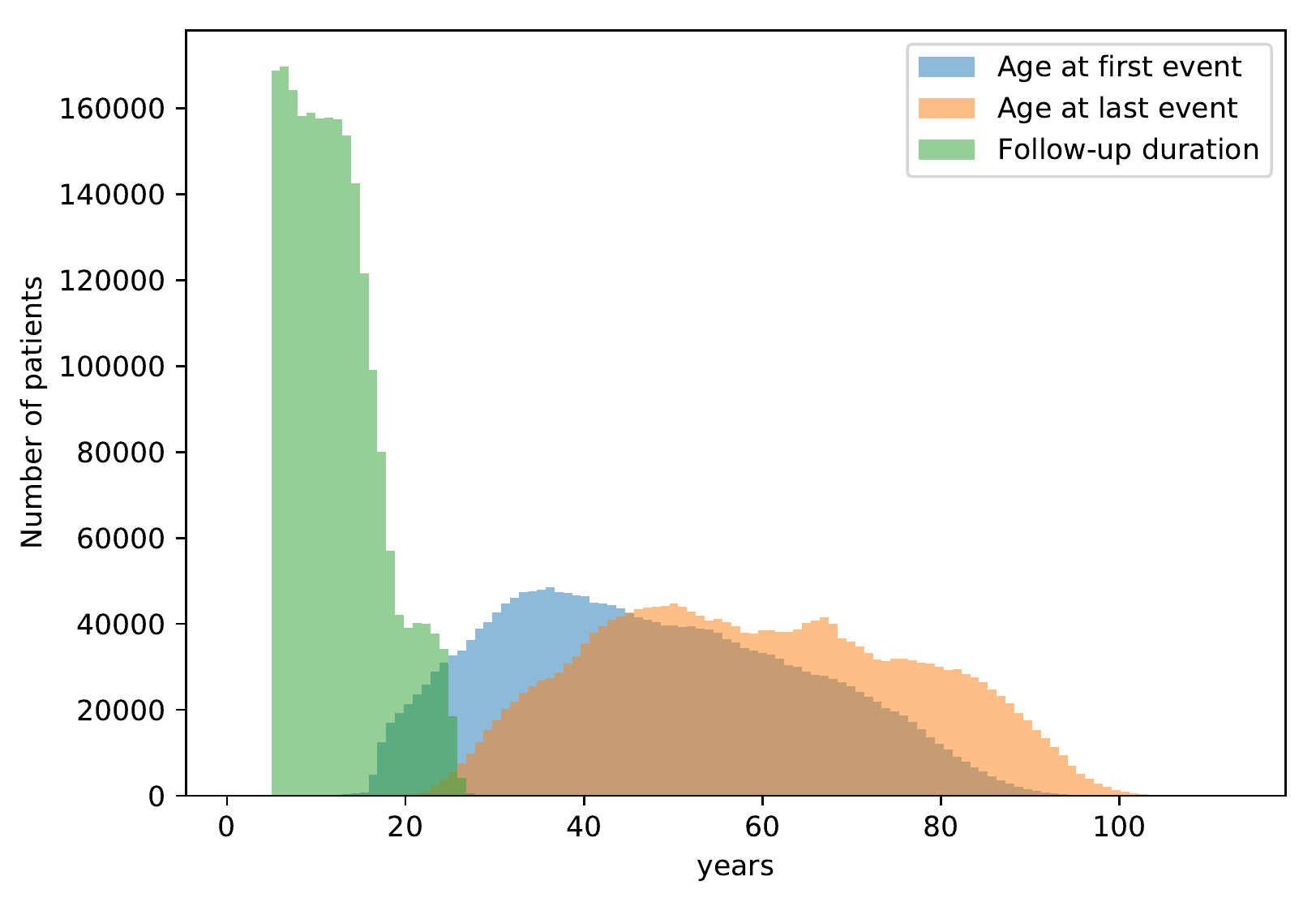}
			\caption{Age distribution and follow-up duration}
	\end{subfigure}	
	\begin{subfigure}{1\linewidth}
		\includegraphics[width=1\linewidth,height=0.3\paperheight,keepaspectratio=false]{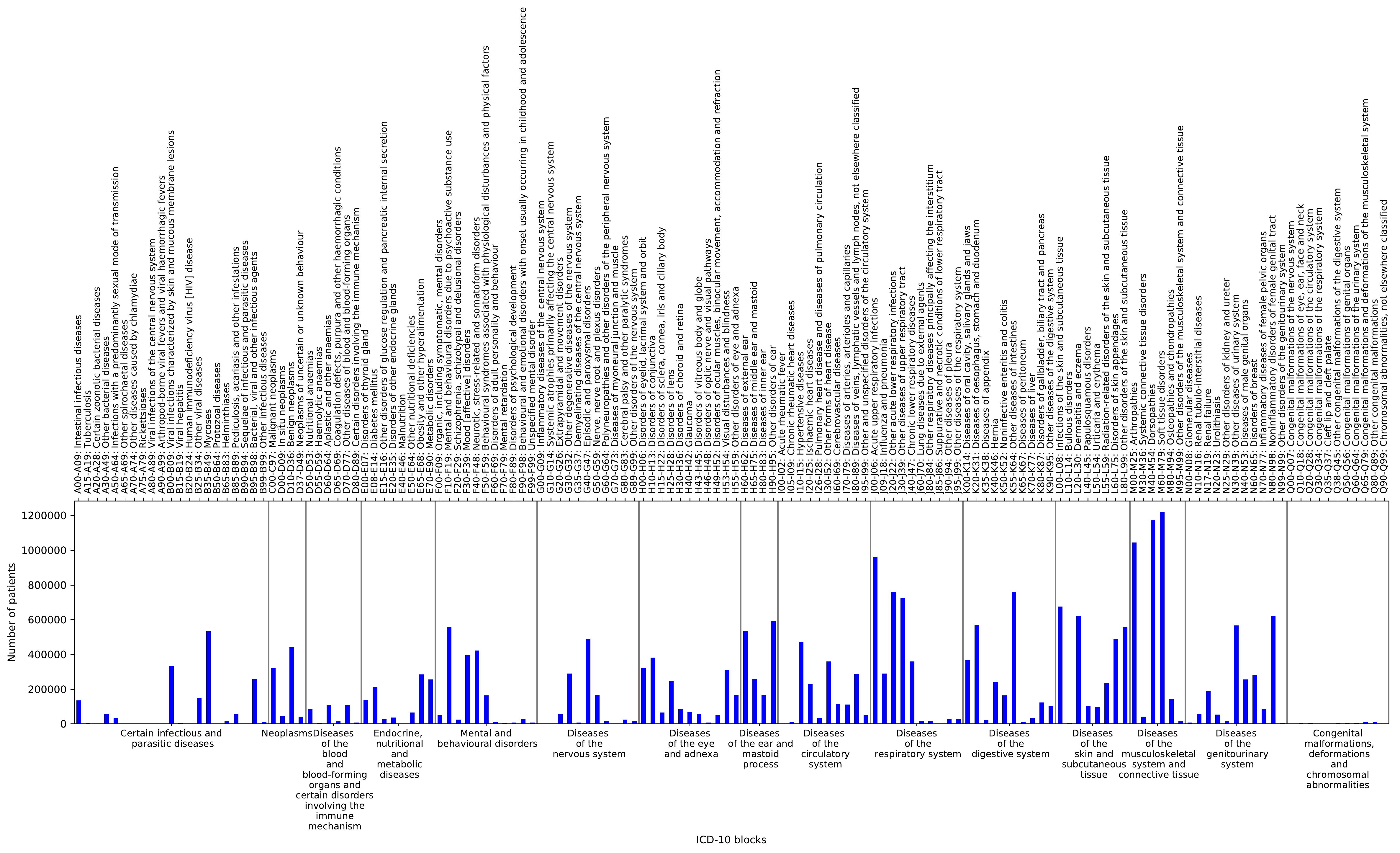}
		\caption{Number of unique patients for each disease}  
	\end{subfigure}	
	\begin{subfigure}{1\linewidth}
		\includegraphics[width=1\linewidth,height=0.2\paperheight,keepaspectratio=true]{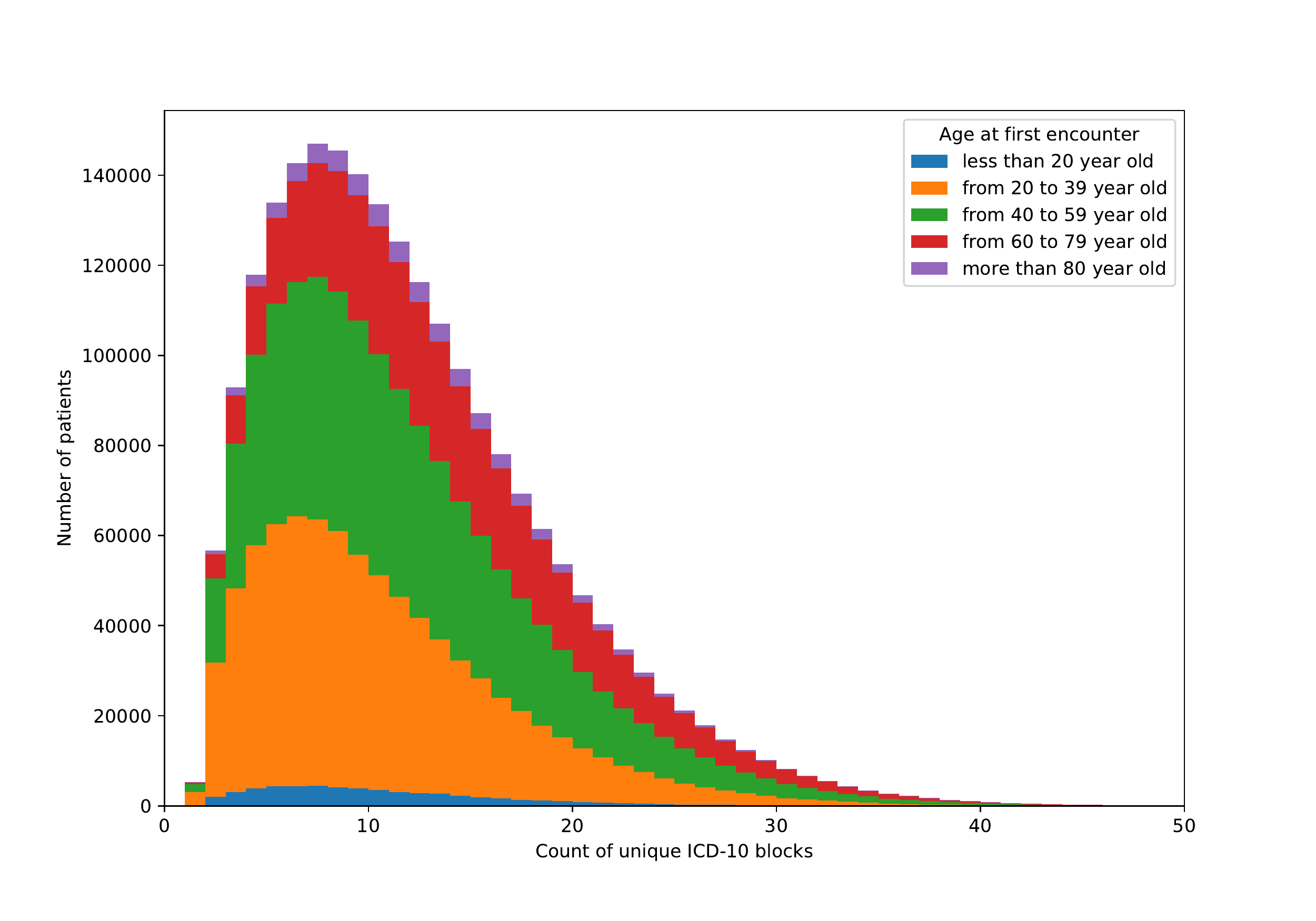}
		\caption{Number of patients for each disease, grouped by age ranges} 
	\end{subfigure}	
	\caption{An exploratory analysis of the data, showing the key characteristics of the population we studied.} 
	\label{fig:exploratory_data_analysis}
\end{figure}

At the end of this process, our cut of CPRD dataset includes nearly 2.2M adult patients (aged 16 years and over) and 65M events; from here, this will be referred to as the data in this study. Figure \ref{fig:exploratory_data_analysis} shows some of the key characteristics of the data. When compared to the past studies, in terms of number of patients, length of follow-up and being representative of the population, our data is comparable to, if not better than, the best-case studied so far.

\subsection{Non-negative Matrix Factorisation}
\label{nmf}
Non-negative matrix factorisation (NMF, or NNMF) refers to a group of algorithms that decompose a matrix $\Dm$ into (usually) two matrices $\Am$ and $\Bm$, with the property that all three matrices have no negative elements, i.e., 

\begin{equation}
\label{eq.nmf}
\Dm\approx \Am\times\Bm\mid\Am\geq0, \Bm\geq0.
\end{equation}

This non-negativity makes the resulting matrices easier to inspect and interpret. Also, in applications such as processing of count data (i.e., the starting point of our multimorbidity analysis), non-negativity is inherent to the data being considered. Since the NMF problem does not have an exact analytical solution in general, there has been a range of numerical approximations for it~\cite{lee2001algorithms,li2001learning,lee1999learning,jia2004fisher,brunet2004metagenes,zhang2007binary,arngren2009bayesian,Zitnik2012,tepper2016compressed,kapralov2016fake}. In this paper, we use the Kullback-Leibler divergence and simple multiplicative updates~\cite{lee2001algorithms,lee1999learning}, enhanced to avoid numerical underflow~\cite{brunet2004metagenes} as implemented by Nimfa python package~\cite{Zitnik2012}.

\subsection{Modelling Pipeline}
We start our modelling pipeline by forming a ``disease matrix'' $\Dm_p$ for each patient $p$, where $\Dm_p(i,j)=1$ if patient $p$ had the first incidence of disease $i$ at age (in years) $j$; $\Dm_p(i,j)=0$ otherwise. This makes $\Dm_p$ a $T\times C$ matrix, where $C$ is the number of conditions/diseases (i.e., 142 in our case) and $T$ is the maximum age we track a patient for (i.e., 114 years in our case). Denoting the total number patients by $N$, this process will result in $N$ such $\Dm_p$ matrices. We carry out two processing steps to create a final disease matrix $\Dm$ for the NMF analysis.

Given the variability in disease prevalences (i.e., some diseases are more common than others), the counts corresponding to rare diseases (such as tuberculosis) are expected to be much lower than the counts corresponding to more common diseases (such as respiratory infections). Therefore, when solving the NMF model, the results can be biased towards explaining the more frequent diseases (i.e., the higher counts). In order to correct for this, in our first processing step, we used an adjustment inspired by TF-IDF (term frequency-inverse document frequency), which is commonly used in natural language processing and information retrieval~\cite{rajaraman2011mining}. More specifically, we introduce DF-IPF (disease frequency - inverse patient frequency), which equals $DF \times IPF$, where we set $DF=1$ as we only considered the first occurrence of each disease. Then for each disease $i$, we defined $IPF(i)=\log(N/N_i)$, where $N$ is the total number of patients and $N_i$ is the number of patients who had disease $i$. The DF-IPF adjustment for $\Dm_p$ will simply result from the multiplication of its entries with the appropriate inverse patient frequency, i.e., $\Dm_p(i,j)\gets \Dm_p(i,j)*IPF(i)$.

As each patient will only have a relatively small number of diseases, $\Dm_p$ is expected to be sparse. On the other hand, NMF does not explicitly model age as a temporal concept (i.e., no explicit model for the relationship among rows in $\Dm$). In medicine, however, one will not see a meaningful difference between a disease happening at age $a$ or $a\pm$1 or 2 years; in factorisation of a matrix like $\Dm$ using standard NMF, this property will not exist and the two scenarios will not be seen as similar. Furthermore, we know from the practice of medicine that the date at which certain chronic disease gets recorded is a noisy concept: For instance, one's hypertension diagnosis can be delayed by months or years due to not noticing or ignoring the symptoms, and/or delaying a doctor visit; or, diseases such dementia are known to have long preclinical periods, where patients who visit their doctors less regularly can, on average, be more likely to have their diagnosis delayed. Therefore, in a second adjustment of $\Dm_p$, and in order to address NMF's lack of temporal regularisation, and take into account the noise and/or uncertainty we have around the age that a particular disease is recorded at, we smooth each column of $\Dm_p$, using a Gaussian filter $g_\sigma$; the optimal value of $\sigma$ will be determined empirically. 

\begin{figure}[H]
\begin{center}
	\includegraphics[width=1.1\linewidth,keepaspectratio=true]{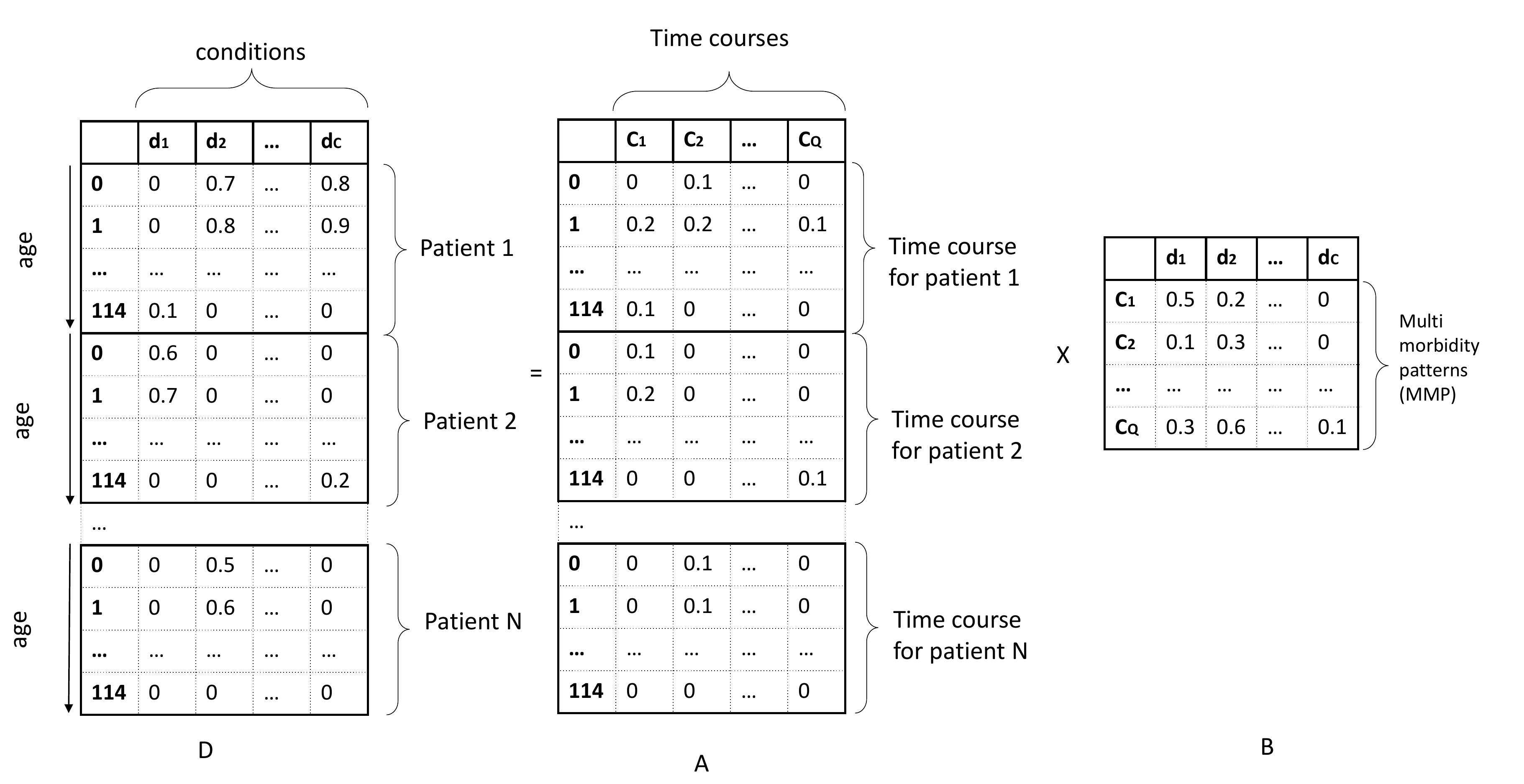}
\end{center}
\caption{An illustration of formation and decomposition of disease matrix $\Dm$. As the figure shows, $\Dm$ results from vertical concatenation of $\Dm_p$ matrices. The matrix $\Bm$ resulting from NMF decomposition has $Q$ rows, each denoting a disease cluster (i.e., a vector of $C$ scores, denoting each disease's belonging to that cluster). The strength of the $q$'th cluster's expression in each patient-age can be found in the $q$'th column of $\Am$, at the entry that corresponds to that patient and age.}
\label{fig:matrix_decomposition}
\end{figure}

As mentioned earlier, to the best of authors' knowledge, the only MP analyses that took the time into account used {\em tensor} factorisation methods. Therefore, one of the key contributions of this paper is to enable {\em matrix} factorisation techniques to result in temporal phenotyping. Assuming that DCs are the same across all patients -- and what varies from patient to patient is the strength of these DCs expressing themselves at different ages -- enables us to use NMF in a unique way that enables temporal phenotyping. 
That is, before we carry out the NMF analysis, we perform one last step: Concatenating the $\Dm_p$ matrices across the age dimension to form $\Dm$ (i.e., $\Dm=[\Dm_1, \Dm_2, ..., \Dm_N]$); a $(T*N)\times C$ matrix. Decomposing $\Dm$ to $Q$ DCs (or factors), makes $\Am$ a $(T*N)\times Q$ matrix and $\Bm$ a $Q\times C$ matrix. Each row in $\Bm$ is a DC (i.e., what we assumed are the same for all patients
%and underly their individual diagnoses
); a column in $\Am$, on the other hand, is the time-course associated with a DC across a number of patients (i.e., how strongly a DC is expressed in a patient at a given age). See Figure~\ref{fig:matrix_decomposition} for an illustration of our NMF analysis, and Table~\ref{tab:glossary} for a glossary of some of the terms introduced in this paper and frequently used in for describing the methodology.

\begin{table}
\caption{Glossary of the terms frequently used in this paper, when describing the methodology.}
\label{tab:glossary}
\begin{tabular}{|p{5.5cm}|p{9cm}|}
\hline
\textbf{Term} & \textbf{Definition} \\\hline
Disease cluster (DC), or multimorbidity clusters (MC) & Each DC (or MC) is a vector of which entries are weights that are assigned to each and every disease to denote the extent to which they belong to it. Note that, as implied, each disease can belong to more than one DC.  \\\hline
DC time course & The strength by which a DC is expressed in a given patient over time. For instance, if the time course shows a larger value at age 50 than 40, it means that its corresponding DC has a stronger expression/presence at age 50 than 40. \\\hline
Multimorbidity pattern (MP) and temporal phenotype & In our study, each MP or temporal phenotype consists of a DC and its corresponding time course. For instance, our NMF approach, decomposes $\Dm$ to a matrix of DCs (i.e., $\Bm$) plus a matrix of their time courses (i.e., $\Am$). An MP might also be referred to as component (e.g., in PCA and ICA) or factor (e.g., in NMF) \\\hline
\end{tabular}
\end{table}

Note that, in this study we focus on disease-based temporal phenotyping, but the approach we introduce can easily incorporate additional phenotypes (e.g., medications) to the input matrix and result in temporal phenotypes that are not disease-only.

\subsection{Ascendency Analysis}
\label{asc.an}
While mining disease clusters is a key objective of multimorbidity research, providing a view into how such clusters evolve over time is another important goal of multimorbidity research that is less studied. Given that our approach results in a time course for each cluster, we aim to treat these clusters as nodes and use their time courses to find their connectivity. More specifically, we employed a simple network modelling technique, which was originally introduced by Patel et al~\cite{patel2006bayesian} for the study of the networks in the brain; it was shown to outperform many network modelling techniques in both finding edges and their directions in a range of settings where ground truth for network edges are available (i.e., a simulation study)~\cite{smith2011network}.

Imagine we have two DCs, with binary time courses $\vb$ and $\wb$ (each T*N dimensional); in our method, if $\vb$ and $\wb$ are active together and inactive together, we consider them associated/connected. This concept is measured by $\kappa_{\vb,\wb}\in[-1,1]$, and can be seen as an undirected edge in a network; it monotonically increases with joint activation of $\vb$ and $\wb$, as their individual activations remain fixed; it monotonically decreases with $\vb$'s activation as $\wb$'s activation and the joint activation remain fixed, or conversely, it monotonically decreases with $\wb$'s activation as $\vb$'s activation and the joint activation remain fixed; it will be 0 when $\vb$ and $\wb$ are statistically independent. More formally, in this approach a bivariate Bernoulli Bayesian model is constructed for the joint activation of each pair of time courses using a multinomial likelihood with a Dirichlet prior. The data to model the joint activation/inactivation probabilities for the two time courses be:
\begin{equation}
\begin{matrix}
z_1=\sum_{NT} I(\vb=1,\wb=1)\\
z_2=\sum_{NT} I(\vb=1,\wb=0)\\
z_3=\sum_{NT} I(\vb=0,\wb=1)\\
z_4=\sum_{NT} I(\vb=0,\wb=0),
\end{matrix}
\end{equation}

The multinomial likelihood of our data takes the form:
\begin{equation}
p(z|\theta) \propto \prod_{i=1}^4\theta_i^{z_i}, 
\end{equation}

where each parameter $\theta$ is defined as
\begin{equation}
\begin{matrix}
\theta_1=P(\vb=1,\wb=1)\\
\theta_2=P(\vb=1,\wb=0)\\
\theta_3=P(\vb=0,\wb=1)\\
\theta_4=P(\vb=0,\wb=0).
\end{matrix}
\end{equation}

\iffalse
In a Bayesian formulation, one can express their prior belief about $\theta$ by defining a Dirichlet prior which takes the form:
\begin{equation}
p(\theta|\alpha) \propto \prod_{i=1}^4\theta_i^{\alpha_i}, 
\end{equation}
where all $\theta_i \geq 0$ and $\sum_{i=1}^{4}\theta_i=1$ (i.e., our posterior model for $\theta$ will be a Dirichlet with parameters $\alpha_i+z_i-1$). 
\fi

$\kappa$ is defined as a ratio, with its numerator measuring the difference between the joint activation probability and the expected joint activation probability under independence, and its denominator forcing it to range from -1 to 1, i.e., 
\begin{equation}
\kappa=\frac{\theta_1-E}{D(max(\theta_1)-E)+(1-D)(E-min(\theta_1)))}
\end{equation}

where $E=(\theta_1+\theta_2)(\theta_1+\theta_3)$, $max(\theta_1)=min(\theta_1+\theta_2,\theta_1+\theta_3)$, $min(\theta_1)=max(0,2\theta_1+\theta_2+\theta_3-1)$ and
\begin{equation}
D=\left\{\begin{matrix}
\frac{\theta_1-E}{2(max(\theta_1)-E)}+0.5\textrm{, if }\theta_1 \geq E
\\
0.5-\frac{\theta_1-E}{2(E-min(\theta_1))}\textrm{, otherwise}.
\end{matrix}\right.
\end{equation}

Given the connectivity between two nodes with $\vb$ and $\wb$ time courses, if $\vb$ exhibits elevated activity for a subset of the period in which $\wb$ exhibits elevated activity, we consider $\wb$  to be ascendant to $\vb$ and vice versa. This is measured by $\tau_{\vb,\wb}\in[-1,1]$, and $\tau_{\vb,\wb}>0$ means that $\vb$ is ascendant to $\wb$, whereas $\tau_{\vb,\wb}<0$ means that $\wb$ is ascendant to $\vb$. More formally:

\begin{equation}
\tau_{\vb,\wb}=\left\{\begin{matrix}
1-\frac{\theta_1+\theta_3}{\theta_1+\theta_2)}\textrm{, if }\theta_2 \geq \theta_3
\\
\frac{\theta_1+\theta_2}{\theta_1+\theta_3}-1\textrm{, otherwise}
\end{matrix}\right.
\end{equation}

\subsection{Benchmarks and Evaluations}
\label{sec:experiments}
It is always challenging to assess the quality of unsupervised/exploratory models, particularly when there is limited access to relevant objective data on their goodness. Therefore, in this study, we aim to provide the field with an objective approach to evaluate the goodness of the DCs that result from multimorbidity research. We used two different sources of objective medical knowledge, each providing a list of ``comorbid'' disease pairs: (1) Jensen et al.~\cite{jensen2014temporal} concluded a list of 1,556 comorbid disease pairs based on the analysis of a large national health dataset, followed by thorough medical due diligence; and (2) Beam et al. \cite{beam2018clinical} studied a large body of medical literature to conclude their list of 113 comorbid condition pairs. Furthermore, through the study of large body of medical literature Beam et al. \cite{beam2018clinical} also concluded a list of 156 causal pairs, where one disease causes the other. Note that, these are the numbers after converting their disease notations to ICD-10 blocks (i.e., the level at which our analysis has been carried out).

Assume that we denote the $q$'th DC as $\qb_k=\{w_{qi}\}_{i=1}^C$ (where $w_{qi}$ is the weight for disease $c_i$ in this cluster), and each disease pair in the look up as $p_{ij}=(c_i, c_j)$. We can define a score $C_L=a/b$, where $b$ is the number of $p_{ij}$'s such that either $c_i$ or $c_j$ belong to the top $L$ diseases in any cluster (i.e., diseases corresponding to the top $L$ weights in a cluster), and $a$ is the number of $p_{ij}$'s such that both $c_i$ and $c_j$ belong to the top $L$ diseases in the same cluster. Using $C_L$ one can evaluate the goodness of DCs. Furthermore, given that the optimal values of our modelling pipeline are expected to lead to better $C_L$, it can be used to guide the grid search for $\sigma$ and $Q$. We defined a similar score based on causal pairs, to assess the goodness of our ascendancy analysis. Assume that we denote a causal pair from Beam et al. as $p_{i\rightarrow j}$ (meaning $c_i$ causes $c_j$), and an ascendancy edge from our analyses as $o_{m\rightarrow n}$ (meaning an arrow from cluster $m$ to $n$). We defined $A_L=a/b$, where $b$ is the number of $p_{i\rightarrow j}$'s that either $i$ or $j$ belong to the top $L$ diseases in either $\qb_m$ or $\qb_n$, and $a$ is the number of $p_{i\rightarrow j}$'s such that diseases $i$ and $j$ belong to the top $L$ diseases in clusters $\qb_m$ and $\qb_n$, respectively. 

In order to check if the $C_L$ and $A_L$ values we observe are higher than what would be expected by chance alone, we sampled from their distribution under the null hypothesis ($\mathcal{H}_0$) that diseases co-occur at random. Under this $\mathcal{H}_0$, we can permute the disease labels in our NMF results (i.e., matrix $\Bm$), and calculate the corresponding $C_L$ and $A_L$. We carried out this label permutation for 1,000 times to create a nonparametric null distribution for $C_L$ and $A_L$, and used this to assess the statistical significance of our observed $A_L$ and $C_L$ values.

\section{Results}
\label{sec:results}
\begin{figure}[H]
	\centering	
	\begin{subfigure}{1\linewidth}
		\includegraphics[width=1\linewidth,height=0.2\paperheight,keepaspectratio=false]{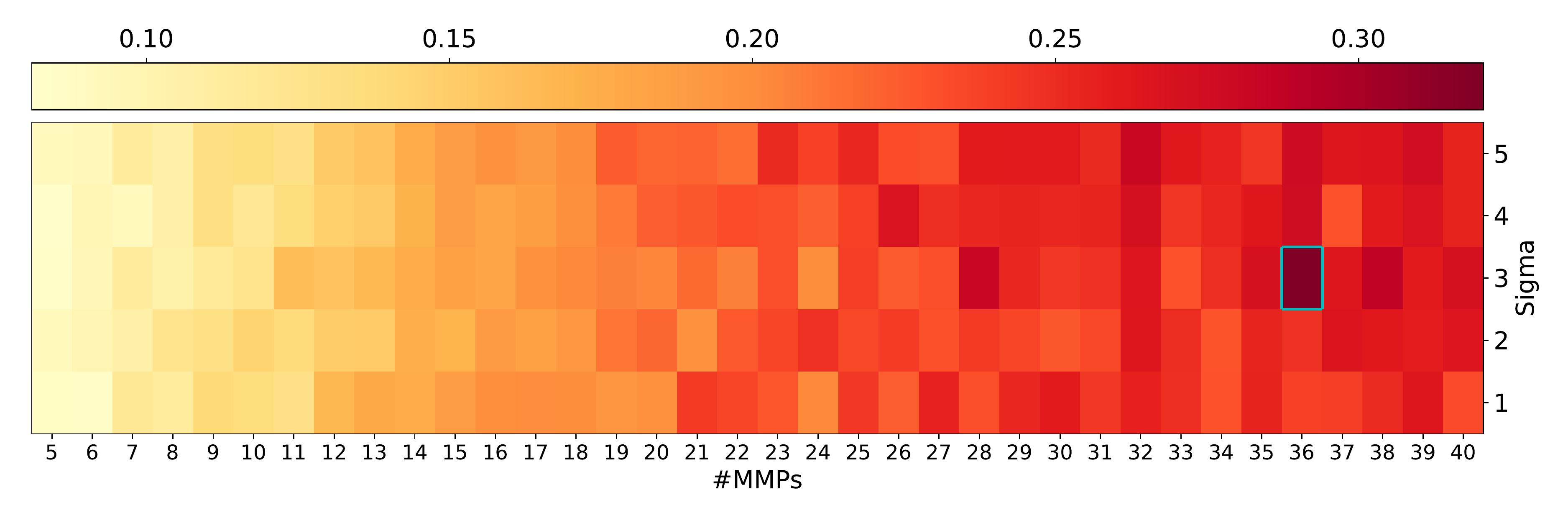}
		\caption{}
	\end{subfigure}	
	\begin{subfigure}{1\linewidth}
		\includegraphics[width=1\linewidth,keepaspectratio=true]{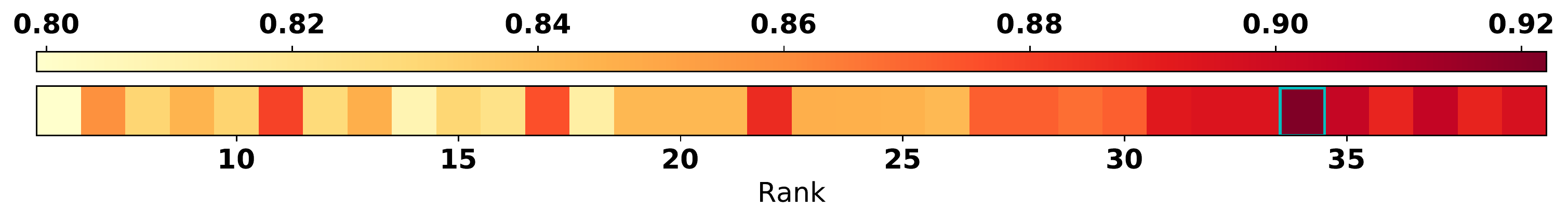}
		\caption{}
	\end{subfigure}	
	\caption{(a)~Evaluation of DCs, using $C_L$, over a 2D grid corresponding to different values of $\sigma$ and $Q$. The highest values (highlighted in cyan border) are observed at $\sigma=3$ and at $Q\approx=36$. All $p$-values are $\leq0.00001$. (b)~The change in cophenetic coefficient as a function of number of factors. As can be seen, 34 seems to be the optimal number of factors for the factorisation of $\Dm$ using NMF.} 
	\label{fig:fine_tuning_q_sigma}
\end{figure}

Our NMF analysis relies on some key free parameters: $Q$ (i.e., the number of DCs) and $\sigma$. We computed $C_L$ for a range of of $Q$ and $\sigma$ values on Jensen disease pairs, which consists of a more comprehensive list of disease pairs than Beam, and hence more suitable for the overall parameter tuning. The resulting $C_L$ scores are shown in Figure~\ref{fig:fine_tuning_q_sigma}.a; using these results, we see a better performance at $\sigma=3$ and $Q\approx36$. In order to find the best value for $Q$, we calculated the cophenetic correlation coefficient, which measures the stability of clustering derived from NMF results at a given rank~\cite{brunet2004metagenes} for a range of $Q$ values. According to the results shown in Figure~\ref{fig:fine_tuning_q_sigma}.b, we see that the factorisation at $Q=34$ results in the highest score. Thus, the rest of our analyses and results will be based on 34 DCs.

Another free parameter of our analysis is the binarisation threshold that we used to prepare the time courses for ascendancy analysis. In order to set this parameter, we calculated $A_L$ for a range of $L$ values at different binarisation thresholds and saw the best results to be at 75\%. Therefore, for the rest of our analysis we thresholded/binarised each time course $\wb$ at $0.75*max(\wb$).

%results are shown in figure \ref{fig:AL_all_thresholds10K}. We can see from the results that for most cases, the optimal value of $P$ is at $75\%$ and will therefore be used hereafter in this study.

%\begin{figure}[H]
%\begin{center}
	%\includegraphics[width=1\linewidth,keepaspectratio=true]{./}
%\end{center}
	%\caption{Evaluation of disease clusters, using $C_L$, over a 2D grid corresponding to different values of $\sigma$ and $Q$. The highest values (highlighted in cyan border) are observed at $\sigma=3$ and at $Q\approx=36$. All $p$-values are $\leq0.00001$.}
	%\label{fig:fine_tuning_q_sigma}
%\end{figure}
%
%
%\begin{figure}[H]
%\begin{center}
	%\includegraphics[width=1\linewidth,keepaspectratio=true]{./cophenetic.pdf}
%\end{center}
	%\caption{}
	%\label{fig:cophenetic}
%\end{figure}

Given that our approach to NMF does not include additional variables such as gender, we decided to carry out the analyses separately for men and women. In our data, we have 994,563 male and 1,209,615 female patients; Figure~\ref{fig:nmf_components} shows the heatmap in which each column corresponds to DCs that result from NMF (i.e., matrix $\Bm$). Note that, corresponding to each one of these clusters we have a time-course that has one value for every patient-age, which shows how that DC is expressed in that patient at that age. %[RK] can you work w Dx or Kazem and put a few of these clusters under the spotlight? Will be great if you add a bit more medical story that goes beyond showing the disease codes that clustered together.

\afterpage{%
\thispagestyle{empty}
\begin{figure}[H]
\begin{center}
\hspace*{-0.8in}
	\includegraphics[height=.7\paperheight,keepaspectratio=true]{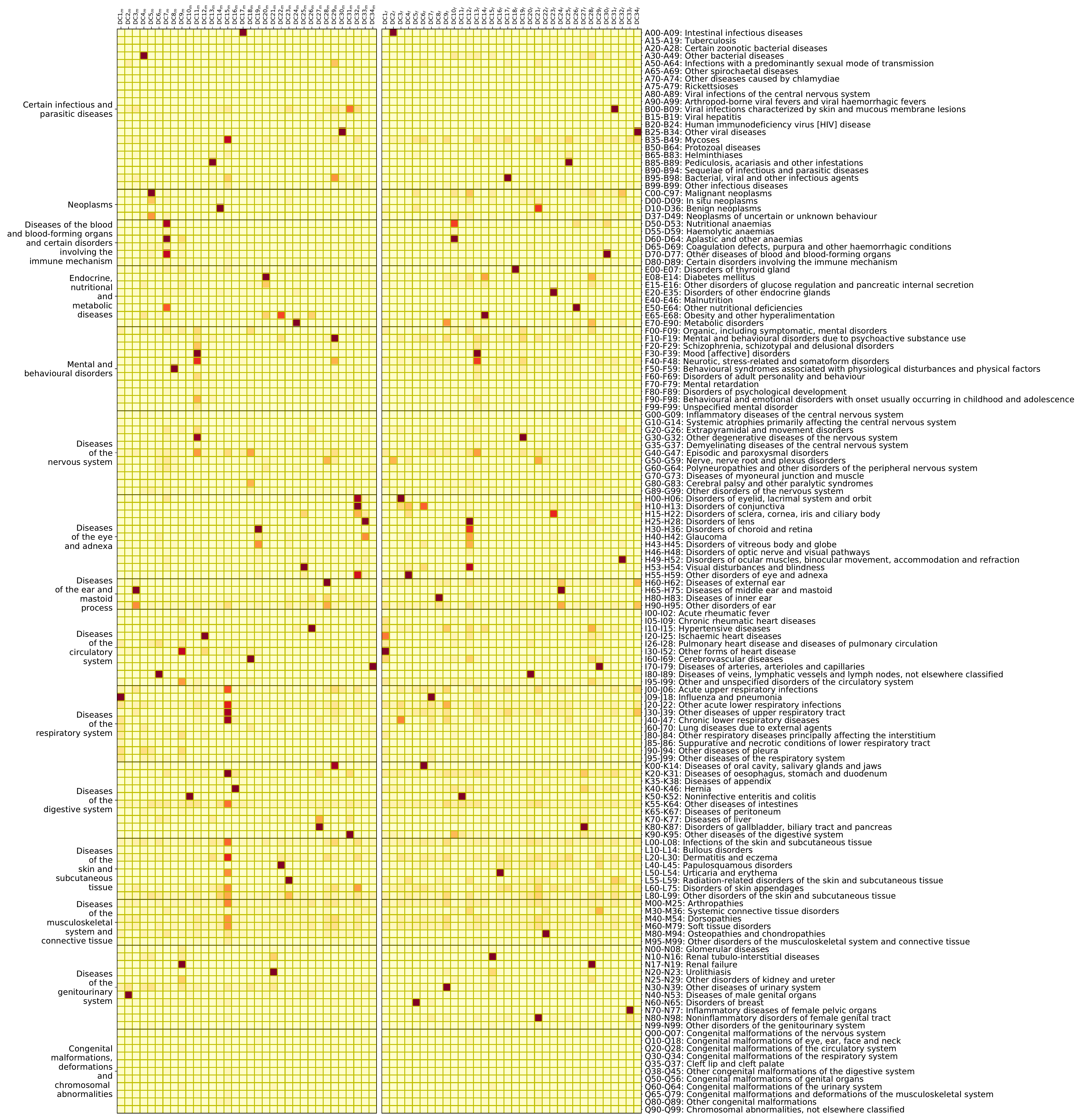}
\end{center}
	\caption{Disease clusters for male and female patients (on the left and right sides, respectively). The figure shows the transposed version ($\Bm^T$) of $\Bm$ matrices, after gamma correction (so that small values are visible).}
	\label{fig:nmf_components}
\end{figure}
}

\begin{figure}[H]
\begin{center}
	\includegraphics[width=1\linewidth,keepaspectratio=true]{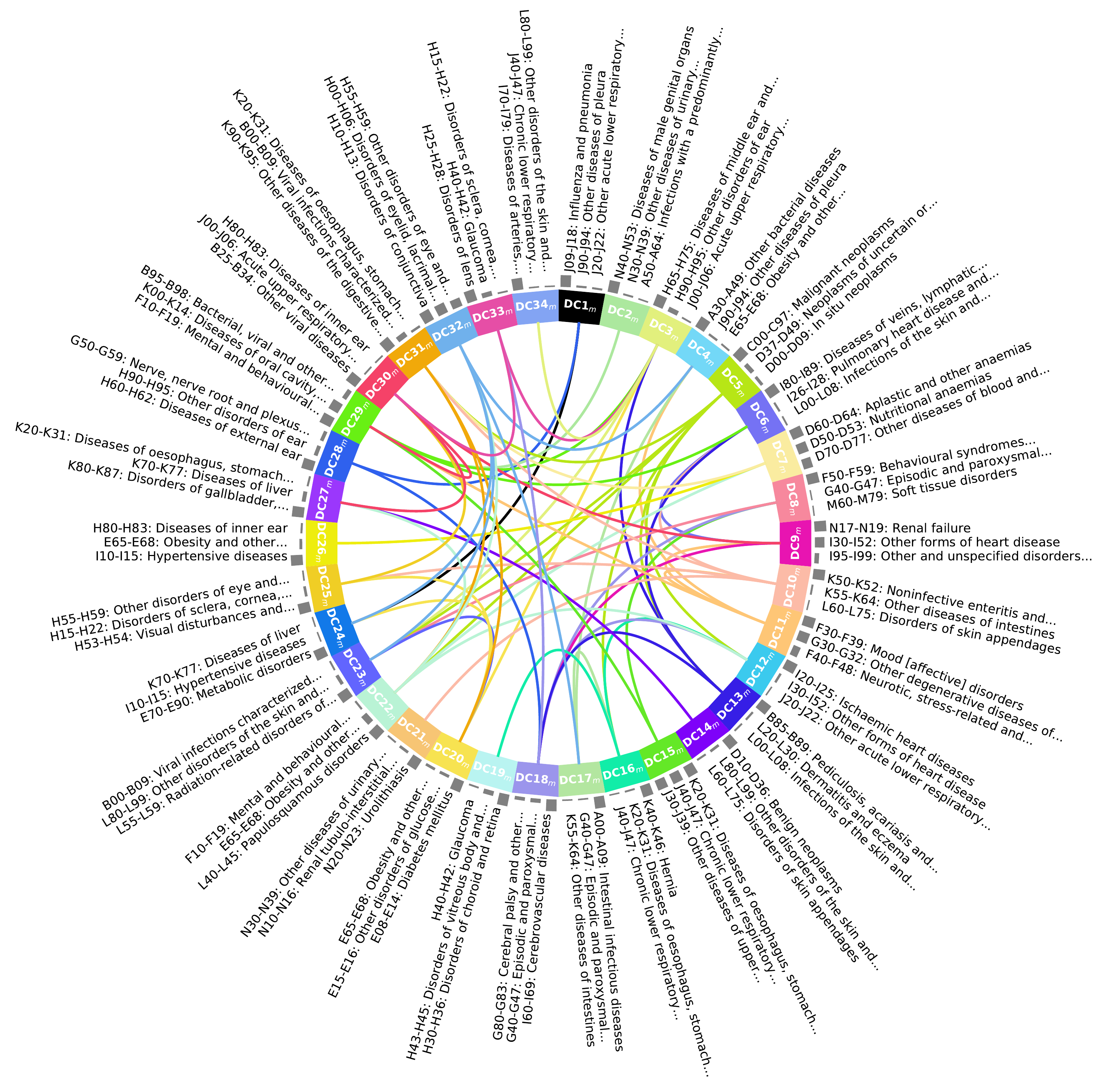}
\end{center}
	\caption{The ascendancy network of MPs for male patients. For visual clarity, we demonstrated the network corresponding to the top 60 edges (i.e., 0.534 threshold for $\kappa$). Note that edges are coloured with the colour of the node they originate from (i.e., the ascendant node).}
	\label{fig:mmp_graph_males}
\end{figure}

\begin{figure}[H]
\begin{center}
	\includegraphics[width=1\linewidth,keepaspectratio=true]{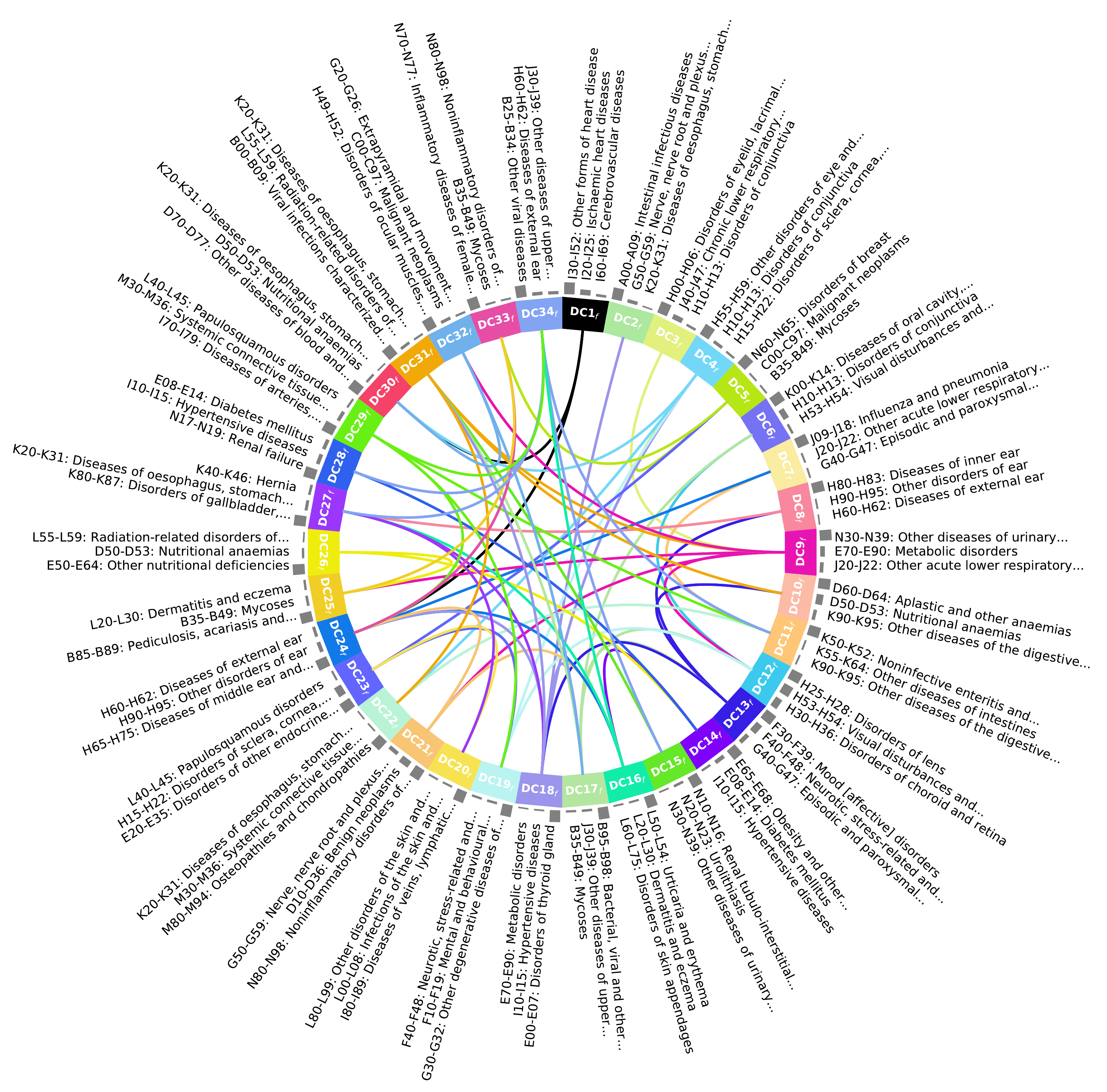}
\end{center}
	\caption{The ascendancy network of MPs for female patients. For visual clarity, we demonstrated the network corresponding to the top 60 edges (i.e., 0.5327 threshold for $\kappa$). Note that edges are coloured with the colour of the node they originate from (i.e., the ascendant node).}
	\label{fig:mmp_graph_females}
\end{figure}

Lastly, to demonstrate how the time-courses that results from our model can be used for the analysis of disease trajectories, we carried out an ascendancy analysis, where each DC is treated as a node and the binarised time courses for each disease pair are used to calculate their corresponding $\kappa$ and $\tau$. The resulting $\kappa$ and $\tau$ can be seen in Figure~\ref{fig:mean_tau_kappa}. Connecting the pairs with high $\kappa$ values, and defining the direction of the resulting edges based on their corresponding $\tau$, a network map can be derived that is shown in Figures~\ref{fig:mmp_graph_males} and~\ref{fig:mmp_graph_females}, for male and female patients, respectively. In both these figures, we chose a threshold for $\kappa$, so that the graph has 60 edges to simplify the visual investigation. Note that for every cluster we show the top 3 diseases (using a bar plot showing their weights). We can see for instance in the network corresponding to male patients (Figure~\ref{fig:mmp_graph_males}) that $DC20_m$ in which the top disease is ``Diabetes mellitus'' leads to $DC24_m$ in which the top disease is ``melabolic disorders''. We can also see in the network corresponding to female patients (Figure~\ref{fig:mmp_graph_females}) that $DC27_f$ in which the top disease is ``Disorders of gallbladder, biliary tract and pancreas'' leads to $DC18_f$ in which the top disease is ``Disorders of thyroid gland''.

\afterpage{%
\thispagestyle{empty}
\begin{figure}[H]
\vspace*{-3cm}
	\centering	
	\begin{subfigure}{.49\linewidth}
	   \includegraphics[width=1\linewidth,keepaspectratio=true]{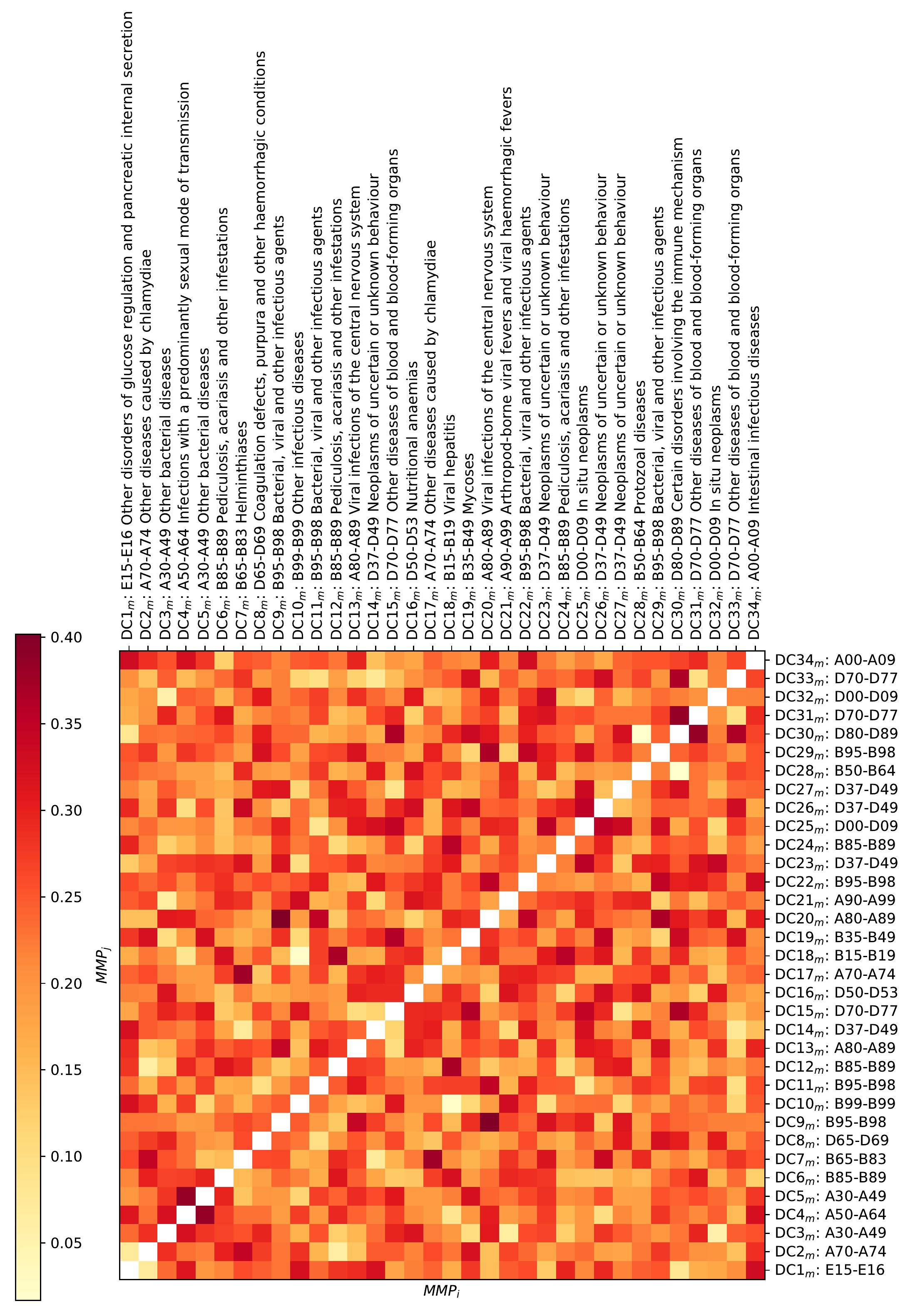}
	   \caption{Mean $\kappa$ for male patients}
	\end{subfigure}	
	\begin{subfigure}{.49\linewidth}
			\includegraphics[width=1\linewidth,keepaspectratio=true]{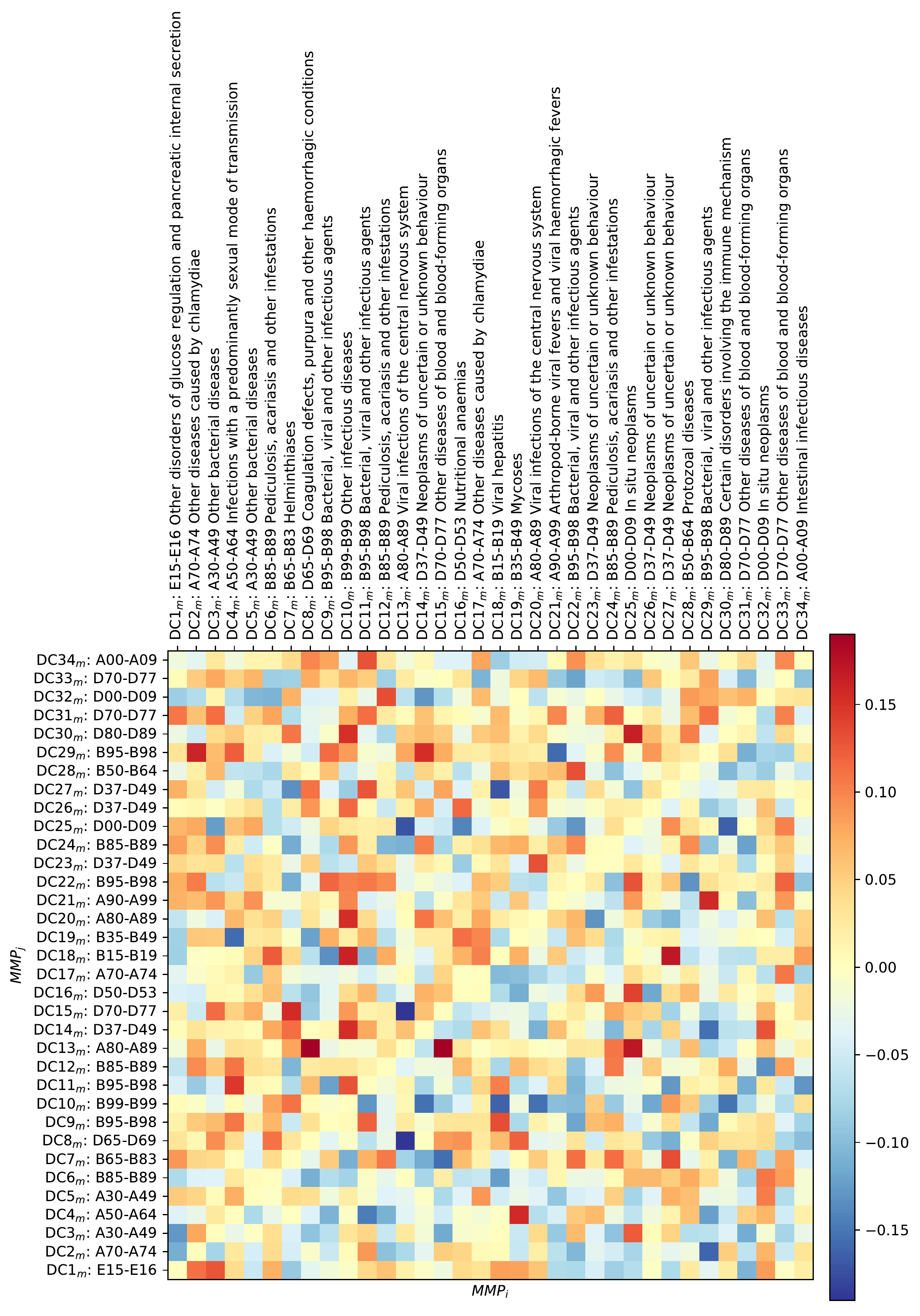}
			\caption{Mean $\tau$ for male patients}
	\end{subfigure}	
	\begin{subfigure}{.49\linewidth}
	   \includegraphics[width=1\linewidth,keepaspectratio=true]{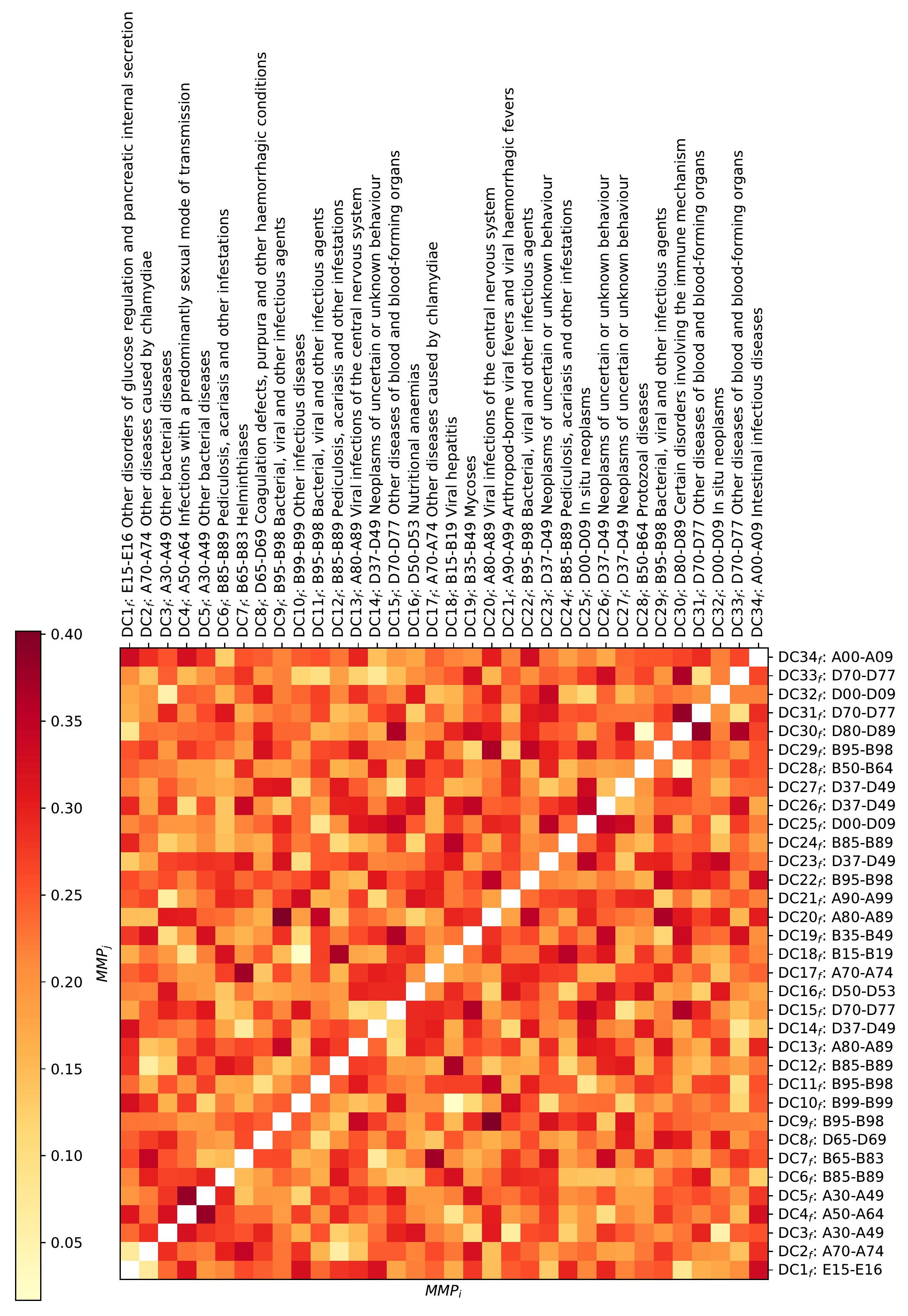}
	   \caption{Mean $\kappa$ for female patients}
	\end{subfigure}	
	\begin{subfigure}{.49\linewidth}
			\includegraphics[width=1\linewidth,keepaspectratio=true]{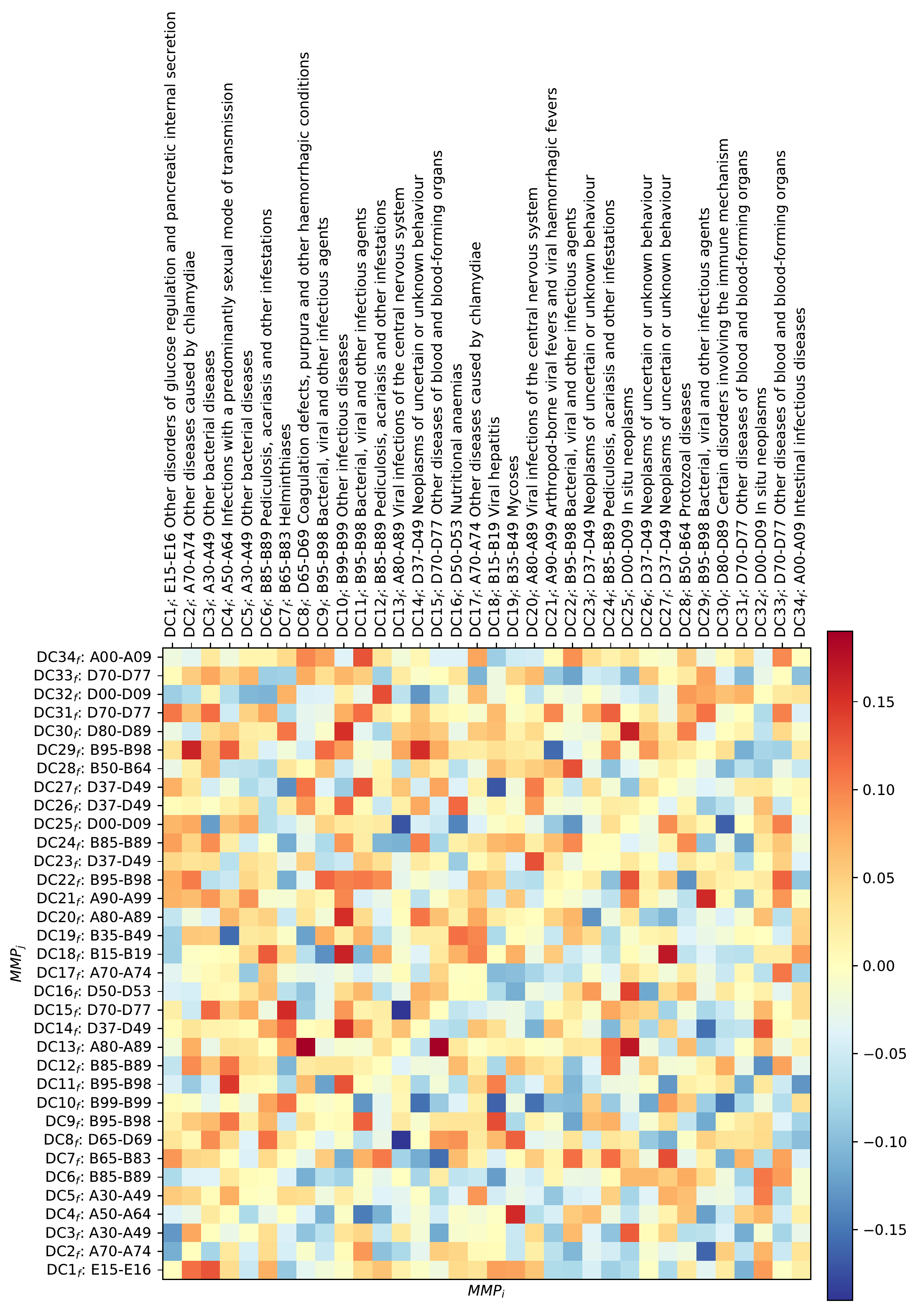}
			\caption{Mean $\tau$ for female patients}
	\end{subfigure}		
	\caption{Mean $\kappa$ and $\tau$ for all pairs of MP components for both male and female patients. Each row/column corresponds to a DC, while the label shows the DC index plus the ICD-10 block with the strongest weight in that cluster. % [RK] can you please modify accordingly? In case you have a curious reader who wants to find the correspondences among figures, if you carry the cluster ID/index from Fig 5 to Fig6-8, it will be very helpful (say, if disease A1-3 are the top three for cluster 10, I'm saying use DC.10: A1, or DC.10: {A1,A2, A3})
	Note that, $\kappa$ is symmetric while $\tau$ is asymmetric (for visual clarity, $\kappa$ values along the diagonal are shown in white to make other values easier to observe).}
	\label{fig:mean_tau_kappa}
\end{figure}
}

%\begin{figure}[htbp]
	%\centering	
	%\begin{subfigure}{1\linewidth}
	   %\includegraphics[width=1\linewidth,keepaspectratio=true]{./network_30nodes_males.pdf}
	   %\caption{MMP ascendancy graph for male patients}
	%\end{subfigure}	
	%\begin{subfigure}{1\linewidth}
			%\includegraphics[width=1\linewidth,keepaspectratio=true]{./network_30nodes_females.pdf}
			%\caption{MMP ascendancy graph for female patients}
	%\end{subfigure}	
	%\caption{Network of MMP components for both male and female patients. The label of each MMP corresponds to the disease with the highest weight.}
	%\label{fig:mmp_graph}
%\end{figure}

%\begin{figure}[ht]
%\begin{center}
	%\includegraphics[width=1\linewidth,keepaspectratio=true]{./network_kappa_larger_05878.pdf}
%\end{center}
	%\caption{Network of MMP components for pairs where $\kappa>0.58$. The label of each MMP corresponds to the disease with the highest weight.}
	%\label{fig:mmp_graph}
%\end{figure}

\section{Conclusions and Discussion}
\label{sec:discussion}

In this study, we employed a well-known matrix factorisation technique called NMF to mine the MPs using one of the largest EHR datasets in the world (i.e., CPRD). The key reason behind this research was to provide a simple and effective solution for multimorbidity research. To be more specific, our study attempted to build on the past studies' learnings, while addressing some of their limitations (e.g., using relatively small data, relying on a narrow observation windows, focusing on a small number of diseases, solely extracting the DCs (instead of temporal phenotyping), and lack of appropriate quantitative benchmarking and evaluation of the results). To the best of the authors' knowledge, our approach is the first one in the literature that has used a matrix (as opposed to tensor) factorisation technique for temporal phenotyping and the study of MPs' temporal patterns.

Enabling NMF to result in temporal phenotyping was achieved through a simple assumption, which is fairly common across similar studies: Every disease belongs to a number of DCs (with a degree of membership); a combination of these clusters for each patient at each year of life explains the observed patterns of diagnoses. This is very similar to a common approach in the study of functional MRI data in neuoimaging known as multi-session ICA through temporal concatenation~\cite{jenkinson2012fsl}.
In cases where one is looking for DCs that are common across population, without assuming them having a consistent expression across different people's lifetime, this approach can be even preferred over the tensor-based factorisation techniques, which assumes the same cluster and time course for everyone. 

Given a number of DCs, the next important objective of multimorbidity analyses is to conclude a disease network, which summarises how diseases interact with one another and influence each other's occurrence. While network models have the potential to solve such a problem (when given the time courses for diseases), the definition of nodes can be a challenge for researchers. That is, if we operate in the ICD-10 universe, we can see scenarios where the network can have anything ranging from 22 nodes (at chapter level) to 10,138 nodes (at level 4). We know from the network modelling literature that the search space for finding the best network is of a super-exponential size on the number of nodes (i.e., $O(n!2^{\binom n2})$)~\cite{robinson1973counting}. This makes the optimisation for learning a network of 10K nodes a huge challenge; both in terms of data (relatively small number of patients, and low prevalence and hence co-occurrence of most diseases at this level) and computing. On the other hand, using ICD-10 chapters, which will result in network with 22 nodes, is likely to lead to results that are hard to be clinically meaningful and interpretable; due to the heterogeneity of the diseases they each contain. Operating at levels such as ICD-10 blocks and 2-digit ICD-10 codes, while not suffering from too many nodes, is still likely to have many highly correlated/co-occurring nodes that might make sense to be combined (particularly, given the data and computing challenges that we face) when learning large networks. 

In this study, we introduced a new concept for the node (i.e., the DCs resulting from the NMF) and a new framework to mine these nodes' relationships with each other. This definition of nodes has a few advantages. Firstly, as our analysis suggests, it leads to a relatively small number of nodes, for which the corresponding network will be easier to learn. Secondly, from an empirical point of view, given that such nodes are driven by diseases that usually co-occur, splitting them into sub-nodes is not likely to be the source of any advantage (specially that such a split will make the network more complex and hence more difficult to learn). And lastly, from a clinical perspective, we are implying that diseases tend to happen in clusters and what the network will tell us is the influence of one cluster on another, given the rest of the clusters. This is in correspondence with what many in the clinical world have been arguing for that the definition of diseases today might not be the most accurate one (and hence various research on phenomapping of the diseases) \cite{goh2007human}.
Based on all these, we carried out our analysis and derived a network, using a simple and yet powerful technique known as ascendancy analysis. Of course, the data from our NMF approach is equally useful for any other network model (e.g., Bayesian networks), and hence there is need for follow-up research on the use of alternative network modelling techniques (an exhaustive list of such techniques have been used and compared in Smith et al~\cite{smith2011network}).

As in most such analyses, our modelling pipeline relied on a mix of choices and assumptions we made; from preprocessing to factorisation technique and beyond. For instance, NMF has a range of different implementations; a follow-up research on comparing different implementations of NMF can surely improve the approach we introduced. Furthermore, the data preprocessing we introduced, from smoothing to adjustment for count can all be done with some differences that can be subject of future research. Lastly, our approach was focused on $\Dm$ as defined in Section~\ref{sec:materials_and_methods}. The earlier works, however, have introduced more concepts (e.g., measurements and medications) to the starting matrix. Given that our approach can simply accommodate additional phenotypes to the input matrix, without necessarily needing to add a new dimension to it (not needing to go from matrix to tensor space) there is an opportunity that can lead to richer analyses that not only can take time/age into account, but also can benefit from additional concepts and interventions that do influence the DCs and their trajectories. However, for the multimorbidity analysis, disease-only analysis was a necessary starting point.

\begin{figure}[ht]
\begin{center}
	\includegraphics[width=1\linewidth,keepaspectratio=true]{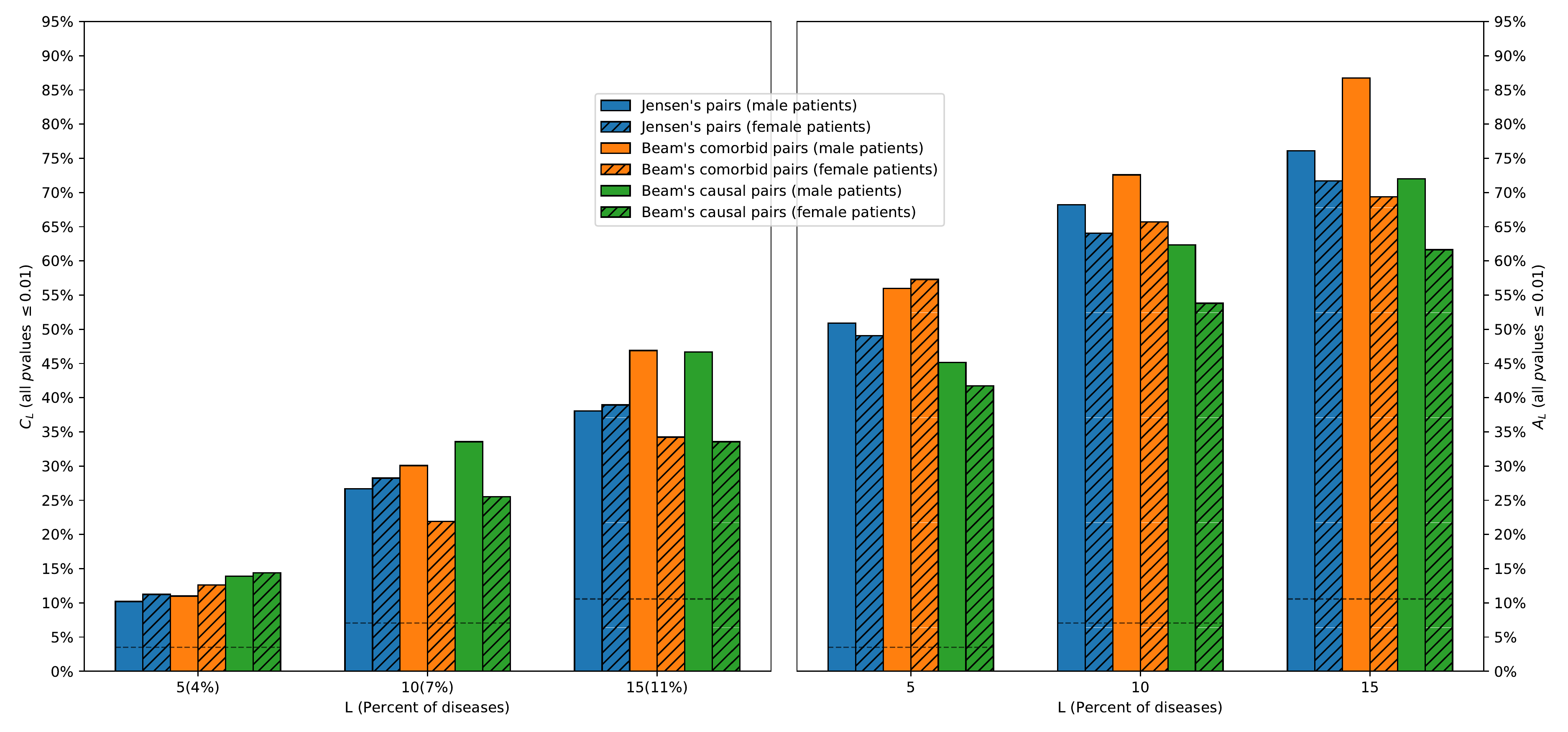}
\end{center}
	\caption{Evaluation of the comorbidity (left) and ascendancy (right). Note that, according to our non-parametric test of significance, all these values/bars have P-values less than 0.001.} 
	\label{fig:AL_CL}
\end{figure}

While we showed some of the results on $C_L$ for parameter tuning earlier, here we also show both $A_L$ and $C_L$ for all the disease pairs, using the optimal values of the free parameters. Note that, while ascendancy analysis does not necessarily mean causal relationship, it has empirically been shown to have high correspondence with it. For instance, Smith et al.~\cite{smith2011network} have shown how ascendancy metrics are better than almost all well-known causal modelling tools, when assessed on simulated data with known ground truth (i.e., simulated data). Furthermore, disease trajectories and various diseases' relationship with each other are complex; there will be cases where different experts might not agree on whether a relationship is causal or comorbid. Therefore, showing $A_L$ and $C_L$ for all pairs will provide additional insight about the strengths and weaknesses of our approach. Figure \ref{fig:AL_CL} shows these results for a number of $L$ values. For instance, at $L=15$, which represents 17\% of our diseases, the results show that our analysis has successfully identified a considerable number of disease pairs in our lookups either as part of the same cluster ($>33\%$), or as linked via ascendancy ($>61\%$).

Lastly, there has been various developments in methods related to our study that can provide multiple new directions for future works. For instance, deep learning's success in the past few years has led to ``deep phenotyping'' research on EHR; while such models can help the study of MPs, their use has been limited to learning disease representations (or embeddings) for disease/event predictions. The earlier works in this space, despite not taking time into account~\cite{tran2015learning,miotto2016deep}, have shown that meaningful DCs can be learned from EHR. Similar results have been shown using CNN~\cite{Nguyen2017Deepr} and RNN~\cite{choi2016retain,rafiq2018deep,xiao2018readmission}. Furthermore, there has been multiple neural solutions for matrix factorisation~\cite{levy2014neural,sainath2013low}. Another novel methodology that has the potential to improve such research is Temporal Regularised Matrix Factorisation (TRMF)~\cite{yu2016temporal}, which has the ability to regularise the temporal aspect of the factors so it is influenced by prior knowledge/assumptions such as smoothness.

\section*{Acknowledgements}
\small{This research was funded by the Oxford Martin School (OMS) and supported by the National Institute for Health Research (NIHR) Oxford Biomedical Research Centre (BRC). The views expressed are those of the authors and not necessarily those of the OMS, the UK National Health Service (NHS), the NIHR or the Department of Health and Social Care.
This work uses data provided by patients and collected by the NHS as part of their care and support and would not have been possible without access to this data. The NIHR recognises and values the role of patient data, securely accessed and stored, both in underpinning and leading to improvements in research and care.}

\bibliographystyle{elsarticlenum} 
\bibliography{bib}




%%\bibitem{}

%%\end{thebibliography}
\end{document}